\documentclass[pdflatex,sn-mathphys-num]{sn-jnl}

\usepackage{graphicx}%
\usepackage{multirow}%
\usepackage{amsmath,amssymb,amsfonts}%
\usepackage{amsthm}%
\usepackage{mathrsfs}%
\usepackage[title]{appendix}%
\usepackage{xcolor}%
\usepackage{textcomp}%
\usepackage{manyfoot}%
\usepackage{booktabs}%
\usepackage{float}
\let\orcidlogo\relax
\usepackage{orcidlink}%

\makeatletter
\renewcommand{\email}[1]{%
  \global\advance\emailcnt by 1\relax
  \if@corauemail
    \g@addto@macro\corrauthemail{\setcounter{footnote}{0}\textcolor{blue}{#1}}%
  \else
    \g@addto@macro\authemail{\setcounter{footnote}{0}\textcolor{blue}{#1}}%
  \fi
}
\makeatother

\makeatletter
\renewcommand{\bmhead}[1]{%
  \par\addvspace{6pt plus 1ex minus .2ex}%
  \noindent{\bmheadfont #1}\par\nobreak\vspace{2pt}%
  \phantomsection\addcontentsline{toc}{subsection}{#1}%
}
\makeatother

\begin{document}
\title[Copewell]{Copewell: A Multi-Agent Swarm Architecture for Equitable Mental Wellness Support}
\author*[1]{\fnm{Seren} \sur{Yenikent}\,\orcidlink{0000-0003-4834-5326}}\email{seren@drcope.health}
\author[1]{\fnm{Jack} \sur{Vinijtrongjit}}
\author[1]{\fnm{Katherine} \sur{Ng}}

\affil*[1]{\orgname{Copewell}, \orgaddress{\country{Singapore}}}

\abstract{Mental health disorders affect nearly one billion people globally, yet 75\% of individuals in low- and middle-income countries receive no treatment due to workforce shortages, cost barriers, and stigma. Current AI-powered wellness solutions predominantly rely on single-mode conversational interfaces that suffer high abandonment rates and fail to provide measurable, immediate relief calibrated to users' dynamic emotional states.

This paper presents Copewell, a novel multi-agent swarm system designed to expand access to mental wellness support through human-centered AI principles. Our architecture introduces three technical innovations: (1) a multi-source assessment framework integrating self-reported, physiological, and contextual data to mitigate algorithmic bias; (2) valence-arousal emotion mapping using Russell's Circumplex Model of Affect to route users to specialized AI agents; and (3) dual-mode intervention delivery combining conversational support with evidence-based sensory wellness protocols.

We examine the sociotechnical design considerations underlying Copewell's development, including a privacy-first architecture, embedded ethical oversight through a dedicated Ethics Supervisor agent, and participatory design informed by mental health practitioners. Early practitioner engagement and beta deployment inform design decisions and identify directions for future empirical evaluation. This work contributes to responsible AI discourse by demonstrating how technical architecture can operationalize equity and safety principles from inception.}


\maketitle

\section{Introduction}\label{introduction}

Mental health disorders represent one of the most significant yet underaddressed challenges in global public health. The World Health Organization (WHO) estimates that approximately one billion people worldwide live with a mental health condition, yet the majority receive no treatment or support \cite{bib1}. This treatment gap is most acute in low- and middle-income countries (LMICs), where an estimated 75\% of individuals experiencing mental health difficulties have no access to care, owing to a combination of professional workforce shortages, economic barriers, and persistent social stigma \cite{bib2,bib3}.

The scarcity of mental health professionals is a structural rather than incidental problem. The global median number of mental health workers is 13.5 per 100,000 population, of whom 43\% are nurses, 22\% psychologists, and 16\% psychiatrists, meaning structured therapeutic care remains concentrated in a small fraction of an already scarce workforce \cite{bib1}. LMICs fall well below this threshold with ratios between 1.1 and 2.4 per 100,000. In high-income countries, while professional access is comparatively better (67.2 per 100,000), demand consistently outpaces supply, placing severe economic strain on health systems and prolonging wait times for care \cite{bib4,bib5}. The practical consequence is that the majority of the global population experiencing mental health difficulties faces either no support or a long delay before receiving it.

Digital technologies, and mobile applications in particular, have been proposed as scalable mechanisms for bridging this gap. The proliferation of mental wellness applications, estimated at over 10,000 on major app platforms, reflects genuine public demand \cite{bib6}. However, the evidence base underpinning most of these tools is thin. A systematic analysis of 278 mental health apps has found that only 16\% had published feasibility or efficacy studies \cite{bib7}. Many rely on single-mode conversational interfaces that provide scripted or narrowly responsive interactions, failing to adapt to users' changing emotional states or to offer effective intervention \cite{bib8,bib9}. This arguably results in high abandonment rates; as among mobile health tools, mental health apps tend to have especially low retention, with very few users continuing to engage beyond the first month \cite{bib10}.

These limitations point to a structural problem that most digital wellness tools have been designed to replicate, rather than rethink, traditional models of support. Artificial Intelligence (AI)-native supplementary care is a different kind of care entirely; one defined by availability, immediacy, and continuity across the care pathway rather than the depth of clinical encounter \cite{bib11}. The responsible design of such tools, therefore, demands not an approximation of therapy, but responsive, real-time engagement that acknowledges emotional state, provides immediate relief, and maintains clear pathways to professional care.

This paper presents Copewell\footnote{https://copewell.ai/}, a multi-agent swarm wellness system designed to address these limitations through human-centered AI principles. Copewell is positioned as a wellness companion and support tool, design philosophy of which begins with the recognition that AI cannot and should not replace human mental health professionals, but can meaningfully supplement care, particularly for populations navigating the financial, geographic, and temporal gaps that characterize access to professional mental health support in much of the world.

The paper makes three primary contributions. First, we describe a novel multi-agent swarm architecture that moves beyond single-agent conversational models, deploying a coordinated ensemble of specialized AI agents each calibrated to distinct emotional states and support needs. Second, we present a multi-source assessment framework that integrates self-reported, physiological, and contextual data to mitigate the algorithmic bias arising from reliance on any single input modality. Third, we examine the sociotechnical design decisions underlying Copewell's development, including privacy-first architecture, an embedded ethical oversight mechanism in the form of a specialized ethics agent, and practitioner-informed participatory design, reflecting on how these choices operationalize responsible AI principles at the architectural level.

This paper is organized as follows. Section 2 reviews relevant literature across digital mental health, multi-agent systems, affective computing, and human-centered AI. Section 3 describes the Copewell system architecture in detail. Section 4 examines key sociotechnical design considerations. Section 5 presents formative insights from practitioner engagement, early beta deployment and safety testing. Section 6 situates this work within broader responsible AI discourse and outlines implications, limitations and future research directions. Section 7 concludes.

\section{Background and Related Work}\label{background-and-related-work}

\subsection{Digital Mental Health Landscape }\label{digital-mental-health-landscape}

The past decade has seen remarkable growth in digital mental health interventions, driven by smartphone proliferation, advances in natural language processing (NLP), and growing public recognition of mental health as a priority health domain. Applications such as Headspace, Wysa and Woebot have achieved significant user adoption, demonstrating genuine demand for accessible wellness support \cite{bib12,bib13,bib14}. The evidence base offers substantive grounds for this momentum.

Digital mental health applications have demonstrated genuine and well-documented value across several dimensions. As accessible, low-barrier, scalable tools, they address structural constraints that traditional clinical settings cannot, e.g., bypassing long waiting times to see a clinician, operating around the clock, and reaching populations for whom professional mental health care remains financially or systemically inaccessible \cite{bib15,bib16,bib17}. Fee-based apps average a fraction of the cost of a single therapy session, making mental health support economically accessible to varying financial abilities such as students, young professionals and communities from low-income countries \cite{bib16,bib18}. Beyond improving access, apps have demonstrated a meaningful capacity to reduce stigma by lowering barriers to help-seeking both on personal (e.g., internalized shame, self-stigma) and cultural (e.g., family reputation concerns in collectivist contexts) levels, that often make in-person treatment feel uncomfortable \cite{bib19,bib20}. Clinically, randomized controlled trials and meta-analyses have shown that mental health apps consistently \emph{do} help reduce depression, anxiety and related distress. Significant symptom reduction has been reported for depression (effect sizes \textasciitilde0.28--0.38) and generalized anxiety (\textasciitilde0.26--0.30) compared with control conditions, confirming real, though small-to-moderate benefits \cite{bib14,bib21,bib22,bib23}. Cognitive behavioral therapy (CBT), mood tracking, chatbot systems and professional guidance have been linked to larger effects and further sustained outcomes \cite{bib23,bib24}. Most durably, apps have proven effective as psychoeducation vehicles, delivering mental health literacy, coping skill instruction, and structured self-management content \cite{bib25,bib26,bib27}. Overall, used not as replacements for professional care but as a complementary layer within it, digital mental health applications occupy a distinct and legitimate role in the support ecosystem.

Despite this demonstrated value, the literature consistently identifies a set of structural limitations that constrain the impact of current digital mental health tools, the most fundamental being the gap between adoption and evidence. Most available apps are developed outside clinical or academic frameworks, with market imperatives driving design decisions more often than therapeutic evidence \cite{bib28}. This has resulted in tools that rely on unvalidated content and generic conversational scripts rather than established psychological frameworks, allowing adoption to significantly outpace validation \cite{bib29,bib30}. Beyond credibility, current tools face significant design limitations. Most deploy uniform, single-modality conversational interfaces that apply the same approach regardless of the user's emotional state, context, or need; thus fail to adapt dynamically to the diversity of human psychological experience \cite{bib31,bib32,bib33}. Safety, ethics, and privacy represent a further cluster of concerns. Crisis management in most apps remains rudimentary, often limited to displaying a static hotline number without structured detection or escalation protocols \cite{bib34,bib35}. Furthermore, ethical safeguards are typically implemented as post-hoc content filters rather than architectural features. Data practices across the app landscape have also attracted significant scrutiny, with privacy protections frequently inadequate for the sensitivity of mental health information \cite{bib36}. Finally, questions of bias and equity remain underaddressed \cite{bib37,bib38}. Tools designed primarily for Western, English-speaking populations risk reproducing existing healthcare inequities at scale, with categorical emotion frameworks and culturally-specific interaction norms limiting transferability to the diverse global populations who stand to benefit most \cite{bib39}. Addressing these limitations requires not incremental improvement to existing designs, but a fundamental rethinking of how AI wellness tools are architected - one that embeds evidence, safety, ethics, and equity as foundational principles rather than features added after the fact \cite{bib40,bib41}.

\subsection{Affective Computing and Emotion Modeling}\label{affective-computing-and-emotion-modeling}

Understanding and responding to emotional states lies at the heart of effective mental health support and increasingly, at the heart of mental health AI design. Affective computing, defined as the capacity of computational systems to recognize, interpret, and respond to human emotional states, has emerged as a foundational discipline for building AI tools that can engage meaningfully with psychological experience \cite{bib42}. Its techniques span emotion recognition, sentiment analysis, and multimodal signal processing, drawing on text, voice, physiological signals, and facial expressions to build a comprehensive picture of a user's affective state \cite{bib43,bib44}. In mental health contexts specifically, affective computing enables systems to move beyond static, one-size-fits-all interactions toward dynamically adaptive support that responds to emotional state in real time, which is a foundational pathway to effective personalized digital mental health intervention \cite{bib8,bib43}.

In digital mental health contexts, the choice of emotion model carries significant consequences for system design. Categorical approaches, such as Ekman's basic emotion framework \cite{bib45}, classify emotion into discrete states such as fear, sadness, or anger, and offer interpretability but sacrifice the nuance required for fine-grained computational analysis and cross-cultural applicability for systems designed to serve diverse global populations \cite{bib46,bib47}. Dimensional models, on the other hand, represent emotional states as points in continuous multidimensional space (e.g. Plutchik's wheel of emotions \cite{bib48}). Thus, they offer a more computationally and culturally robust alternative due to enabling richer mathematical processing. Russell's Circumplex Model of Affect \cite{bib49} positions emotional states within a two-dimensional space defined by valence (pleasant--unpleasant) and arousal (high--low activation). This framework has been widely adopted in affective computing and emotional AI modeling as well as has demonstrated cross-cultural consistency \cite{bib46}. The model enables calculation of emotional distances, centroids, and variability metrics; and it translates naturally into mathematical representations suitable for real-time computational processing \cite{bib50,bib51}.

In mental health AI specifically, the valence-arousal framework enables fine-grained differentiation between emotional states that categorical models would treat as equivalent - distinguishing, for example, between anxious high-arousal distress and depleted low-arousal low mood, two presentations that require fundamentally different support responses. Copewell adopts this framework as the computational substrate for its agent routing mechanism, building on an established body of affective computing research to enable emotionally responsive and contextually appropriate support delivery.

\subsection{Multi-Agent Systems in Mental Healthcare AI}\label{multi-agent-systems-in-mental-healthcare-ai}

Single-agent conversational systems have made meaningful contributions to accessible mental health support, yet their architectural uniformity imposes inherent limits on the range of support they can provide. The psychological diversity of mental health manifestations, spanning acute anxiety, low mood, emotional dysregulation, and contemplative states, among others, calls for correspondingly differentiated support orientations that a single conversational mode is structurally limited to address \cite{bib16,bib40}. Research on therapeutic modality matching suggests that the effectiveness of a support approach is partly a function of its fit with the user's current emotional and cognitive state \cite{bib52}, a principle that single-agent architectures cannot readily operationalize at the design level.

Multi-agent systems (MAS) offer a principled architectural response to this constraint. In MAS architectures, autonomous agents, each with distinct capabilities, knowledge, and behavioral orientations, collaborate dynamically to address problems that exceed the capacity of any single agent \cite{bib53}. The core advantage of this approach is specialization: rather than requiring one agent to approximate all support modalities, a swarm of specialized agents can each be optimized for a specific context, with coordination mechanisms determining which agent engages at any given moment.

In healthcare AI, multi-agent architectures have demonstrated value across clinical decision support, treatment planning, and patient monitoring, domains where the complexity of human health states similarly exceeds single-agent capacity \cite{bib54}. In mental health specifically, prior work has explored agentic frameworks for therapeutic support, including RAG-augmented systems that ground conversational agents in verified clinical knowledge \cite{bib55}. \cite{bib56} developed a system using multiple Large Language Model (LLM) agents that debate, then create a tailored counselor persona and generate responses aligned with user preferences and professional standards. \cite{bib57} demonstrated the feasibility of structured multi-agent workflows in mental health contexts, with specialized agents for exploration, therapeutic delivery, and consolidation, with dynamic routing across evidence-based modalities including CBT and Mindfulness-Based Cognitive Therapy (MBCT).

MAS architectures apply the therapeutic modality matching at a computational level by continuously sensing emotional state, explicitly modeling its evolution, and selecting from a repertoire of strategies or agents whose behavior is tailored to that current state and goal \cite{bib53,bib58}. Rather than selecting a therapeutic orientation once at design time and applying it uniformly, a coordinated agent swarm dynamically assigns the most contextually appropriate support mode e.g., de-escalation for acute distress, gentle activation for low mood, reflective support for contemplative states \cite{bib55}.

Copewell extends this paradigm through a swarm architecture in which agent selection is driven by continuous valence-arousal assessment, as described in Section 2.2, rather than keyword matching or intent classification. This integration of affective computing theory with multi-agent coordination represents the architectural contribution at the core of this paper (See Section 3).

\subsection{Human-Centered AI and Responsible Design}\label{human-centered-ai-and-responsible-design}

Mental health AI occupies a uniquely high-stakes design space. Unlike general-purpose AI applications, systems designed to engage with psychological distress operate at the intersection of user vulnerability, sensitive personal data, and consequential support decisions. This makes responsible design not an aspirational quality but a foundational requirement \cite{bib36,bib41}. Design failures in this domain carry direct risks to user safety, dignity, and autonomy that demand a level of ethical rigour commensurate with the sensitivity of the context \cite{bib37}.

Established responsible AI frameworks such as the IEEE Ethically Aligned Design principles, the EU AI Act, and the OECD Principles on AI converge on a core set of requirements: transparency, accountability, fairness, and safety \cite{bib59,bib60,bib61}. In mental health contexts specifically, these abstract principles translate into concrete architectural demands: crisis detection and escalation mechanisms, privacy-preserving data handling, culturally inclusive interaction models, and honest delineation of system scope and limitations. The Framework for AI Tool Assessment in Mental Health (FAITA-MH) operationalizes these requirements into six evaluative dimensions: Credibility, User Experience, User Agency, Equity and Inclusivity, Transparency, and Crisis Management, providing the most directly relevant evaluation lens for AI-powered wellness tools \cite{bib62}.

Responsible AI in sensitive domains further requires that ethical commitments be embedded architecturally rather than applied as post-hoc compliance mechanisms, which is referred to as \emph{ethics by design} \cite{bib63}. For mental health AI, this distinction is particularly consequential since a system that monitors for ethical violations after generating a response offers fundamentally weaker protection than one in which ethical oversight is a structural participant in the interaction.

Finally, responsible design in mental health AI requires meaningful engagement with practitioners and users throughout the development process, not as validation subjects but as co-designers whose domain expertise shapes architectural decisions \cite{bib64}. The nature and implications of this engagement in Copewell's development are discussed in detail in Section 5. Together, these principles constitute not a checklist applied to Copewell's design but the philosophical foundation from which its architecture was derived - each traceable to specific design decisions.

\section{System Architecture}\label{system-architecture}

\subsection{Overview}\label{overview}

Copewell's architecture is organized as a multi-stage pipeline: assessment, routing, and intervention (see Fig.~\ref{fig1}). At the assessment stage, data from multiple sources is integrated to establish a representation of the user's current emotional and contextual state. This representation informs a routing decision that assigns the user to the specialized AI agent best suited to their needs. The selected agent delivers support through a combination of conversational interaction and, where appropriate, evidence-based sensory wellness interventions. An Ethics Supervisor agent operates as a cross-cutting oversight mechanism throughout this pipeline, capable of overriding routing decisions or intervening in conversations when safety protocols are triggered.

This pipeline architecture embeds a set of design commitments: that assessment should be multimodal rather than relying on any single data source; that support should be dynamically calibrated to emotional state rather than uniformly applied; that intervention should address both cognitive-emotional and physiological dimensions of wellbeing; and that safety and ethical oversight should be structural features of the system rather than afterthoughts.

\subsection{Multi-Source Assessment Framework}\label{multi-source-assessment-framework}

A central limitation of single-modality assessment in wellness applications is its susceptibility to systematic bias. Self-reported emotional states are subject to social desirability effects, limited introspective accuracy, and the inherent difficulty of communicating subjective experience. Reliance on self-report alone risks misclassifying a user's actual need, potentially routing them to an inappropriate support modality. Conversely, reliance on physiological signals alone ignores the subjective, experiential dimension of emotional states that self-report captures \cite{bib65,bib66}.

Copewell addresses this through a multi-source assessment framework integrating three data streams: self-reported state (weighted at 40\%), physiological and biometric data (35\%), and environmental and contextual data (25\%). The relative weighting reflects the hierarchy of signal proximity to the user's experienced state: self-report provides direct access to subjective experience, physiology provides objective correlates of arousal, and context provides the situational frame for interpreting these signals. This weighting schema is an original design decision informed by established principles in Ecological Momentary Assessment \cite{bib67} and contextual behavioral science \cite{bib68}.

Self-reported data is collected through structured daily mood check-ins and conversational input from chatbot dialogues. Beyond explicit reports, the system analyses implicit patterns in linguistic expression, including vocabulary breadth, sentence length, and affective tone, to surface signals that may not be consciously reported by the user \cite{bib69}.

Physiological data integration, where available through device health application programming interfaces, includes sleep quality metrics, heart rate variability (HRV), and activity levels. These biometric signals are meant to be used for proxy as objective correlates of arousal and stress that complement and cross-validate self-report.

Contextual data is derived from calendar integration and enables the system to account for situational factors, e.g., a high-meeting week, workout session, regular doctor visit, that provide interpretive context for current state assessment. This layer operationalizes insights from contextual behavioral science, which demonstrates that emotional experience is substantially shaped by environmental circumstances that individuals may not consciously connect to their current state \cite{bib68,bib70}.

The cross-validation of these three data streams is designed to prevent the category of algorithmic error that arises when a system takes an isolated signal as definitive. A user who reports feeling well but whose sleep data indicates severe disruption and whose calendar shows sustained overload represents a different risk profile from one in whom all three signals are aligned.

\subsection{Valence-Arousal Emotion Mapping}
Following assessment, the integrated data is used to position the user's current state within the valence-arousal space defined by Russell's Circumplex Model \cite{bib49}. This mapping serves two functions: it provides a computationally tractable representation of the emotional state that supports routing decisions, and it enables tracking of emotional trajectories over time, informing the predictive analytics layer described in Section 3.7.

The valence dimension captures the pleasantness (e.g., happy, serene) or unpleasantness (e.g., upset, lethargic) of the current emotional experience. The arousal dimension captures the activation level, ranging from high-energy states (e.g., excited, tense) to low-energy states (e.g., calm, fatigued). Together these dimensions define four functional quadrants, each associated with meaningfully distinct support needs and corresponding to a specialized AI agent.

The mathematical advantages of this continuous representation include the ability to calculate emotional distances (i.e., how far a user's state is from a neutral or positive baseline), centroids (i.e., the typical emotional position across a session or period), and variability scores (i.e., the degree of emotional fluctuation, which may itself carry interpretive significance) \cite{bib50,bib51}. These metrics aim to enable quantitative characterization of a user's emotional pattern over time and provide a foundation for both personalization and longitudinal pattern analysis within the system's wellness scope.

\subsection{Agent Routing and Specialization}\label{agent-routing-and-specialization}

The four quadrants of the valence-arousal space are associated with four specialized AI agents, each calibrated to the support needs characteristic of that quadrant (see Fig.~\ref{fig2}):

\begin{itemize}
\item
  Quadrant 1 (High Arousal, Positive Valence - e.g., energized, excited, motivated): This quadrant captures states of high physiological and psychological activation paired with positive affect. Users in this state experience elevated energy levels alongside pleasant emotional valence. The \emph{Facilitator Agent} serves this state by channeling productive energy through structured activities, goal-setting interactions, and forward-oriented conversation.
\item
  Quadrant 2 (High Arousal, Negative Valence - e.g., anxious, stressed, angry): This quadrant captures states of high activation paired with negative affect and users in this state experience heightened arousal alongside unpleasant emotional valence. The \emph{Stabilising Agent} serves this state by prioritizing de-escalation and grounding, with a conversational approach emphasizing validation, pacing, and the provision of immediate calming interventions. This agent is most likely to initiate handoff to sensory wellness modalities such as breathwork guidance or visual grounding content (See Section 3.6).
\item
  Quadrant 3 (Low Arousal, Negative Valence - e.g., fatigued, disengaged, low mood): This quadrant captures states of low activation paired with negative affect with users experiencing diminished energy alongside unpleasant emotional valence. The \emph{Motivator Agent} serves this state through gentle activation strategies, hope-building, and behavioral activation principles informed by cognitive-behavioral frameworks. Its conversational pacing is calibrated to the user's current energy rather than demanding high engagement.
\item
  Quadrant 4 (Low Arousal, Positive Valence - e.g., calm, content, reflective): This quadrant captures states of low activation paired with positive affect. Users in this state experience reduced arousal alongside pleasant emotional valence. The \emph{Reflector Agent} serves this state by supporting contemplation, meaning-making, and insight development, that is suited to journaling support, gratitude practice, and mindfulness-oriented interaction.
\end{itemize}

This quadrant-based routing system embeds a theoretically grounded principle that the effectiveness of a support approach depends on its fit with the user's current state directly into the architectural design \cite{bib52}. Rather than applying a uniform conversational approach regardless of context, Copewell dynamically selects the agent whose orientation is most likely to be beneficial given the assessed emotional state. Routing is not static within a conversation. Real-time linguistic analysis monitors shifts in expressed emotional state, enabling handoff between agents when a meaningful transition is detected. Transitions are designed to preserve narrative coherence; thus, the user experiences a continuous conversation rather than a perceptible agent switch, achieved through shared context retention and continuity protocols.

\subsection{The Ethics Supervisor Agent}\label{the-ethics-supervisor-agent}

A distinctive architectural feature of Copewell is the Ethics Supervisor agent, which is a dedicated oversight mechanism that operates as a cross-cutting layer across all agent interactions. Unlike approaches that implement ethical safeguards as static system checkpoints or post-hoc content filters \cite{bib40,bib63}, Copewell's Ethics Supervisor is an active participant in the agent swarm, capable of observing, evaluating, and intervening in real time.

The conceptual basis for this design draws an analogy to regulatory network functioning in human cognition: prefrontal cortex systems are often described as supporting the evaluation and regulation of more automatic, affect-driven processes \cite{bib71}. In Copewell's architecture, the Ethics Supervisor performs an analogous function monitoring the outputs of specialized agents for alignment with ethical guidelines, safety protocols, and scope boundaries, and intervening when these are threatened.

The Ethics Supervisor's intervention authorities span two distinct domains. In the domain of ethical oversight, it flags responses that make implicit clinical claims beyond Copewell's wellness scope, redirects conversations approaching clinical boundaries (e.g., discussions that move from emotional support toward diagnostic framing), and ensures data handling aligns with privacy commitments. In the domain of safety, it monitors all conversational interactions for acute risk indicators (e.g., linguistic markers associated with suicidal ideation and self-harm). When these are detected, it overrides all active agents entirely, delivering a validated, empathetic crisis response alongside clear signposting to emergency resources. This dual function is architecturally unified in a single agent because safety and ethics are not separable concerns in a mental wellness context, and they are rather two expressions of the same foundational commitment to user wellbeing.

\subsection{Sensory Wellness Modalities}\label{sensory-wellness-modalities}

A defining architectural choice in Copewell is the integration of evidence-based sensory wellness modalities alongside conversational support. Existing mental health AI systems are predominantly conversation-only, offering no mechanism for direct physiological intervention \cite{bib40}. This single-channel design limits the system's capacity to deliver measurable and immediate relief, particularly in high-arousal states where cognitive engagement alone may be insufficient or contraindicated. Copewell employs a sensory wellness library targeting physiological and perceptual mechanisms of regulation.

The sensory library is organized across audio, visual, and combined multi-sensory components, delivered through a three-tier system that accommodates user preferences and environmental contexts. The audio-only tier supports use in contexts where visual engagement is impractical, e.g., eyes-closed practice, commuting, low-bandwidth environments, drawing on binaural beats, Autonomous Sensory Meridian Response (ASMR) content, and therapeutic frequency delivery. The visual-only tier supports users with sound sensitivity, those in shared environments, or those who respond more strongly to visual processing, using fractal patterns, color therapy, and silent brainwave entrainment. The full multi-sensory tier coordinates synchronized audio and visual delivery for maximum therapeutic engagement when context permits.

Each modality draws on a distinct body of research linking sensory input to measurable physiological and affective outcomes. ASMR has been associated with reductions in heart rate and improvements in mood across controlled conditions \cite{bib72,bib73}. Visual entrainment techniques, including brainwave-synchronized flicker patterns at specific frequencies, have been investigated for their capacity to modulate attentional and arousal states across anxiety, low mood, and focus presentations \cite{bib74,bib75}. Fractal pattern exposure, especially at mid-range fractal dimensions, has been associated with preference and some stress-reduction or relaxation-related effects, often explained through fractal fluency \cite{bib76,bib77}. Color therapy interventions, calibrated to wavelength-specific psychophysiological associations, have shown effects on anxiety and mood states in controlled studies, though the evidence base across color modalities varies in strength \cite{bib78}. Brainwave entrainment using binaural beats targets specific frequency bands: alpha (8--12 Hz) for anxiety and stress reduction, gamma (40 Hz) for focus and cognitive performance, delta (0.5--4 Hz) for sleep onset, and beta (13--30 Hz) for mood activation \cite{bib79,bib80,bib81,bib82}. The evidence base across these modalities is variable in strength, and Copewell's integration of them is framed as evidence-informed rather than clinically validated, with ongoing empirical evaluation identified as a priority.

Sensory modality selection is mapped to the valence-arousal quadrant assessed during the routing process described in Section 3.4. Users in Quadrant 2 (high arousal, negative valence) may be offered alpha wave entrainment with blue color therapy and calming fractal patterns - combinations theoretically suited to parasympathetic activation. Users in Quadrant 3 (low arousal, negative valence) may be offered beta wave activation with yellow color therapy and energizing visual patterns..

\subsection{Predictive Analytics and Behavioral Modeling}\label{predictive-analytics-and-behavioral-modeling}

Copewell incorporates a predictive analytics layer that builds longitudinal models of individual user patterns. This layer employs trend analysis, volatility detection, anomaly identification, and cyclical pattern recognition to surface insights about a user's wellbeing trajectory over time.

The predictive layer serves two functions. First, it enables proactive support suggestions when patterns suggest deterioration before the user actively signals distress, operationalizing a preventive orientation. Second, it provides a data foundation for the practitioner-facing dashboard (see Section 6.3 for \emph{Therapist Mode}, currently in development), which gives clinicians visibility into their clients' between-session patterns, supporting continuity of care without requiring additional appointment time.

\section{Sociotechnical Design Considerations}\label{sociotechnical-design-considerations}

This section documents the key sociotechnical design considerations that shaped Copewell's development, spanning data privacy, ethical oversight and crisis safety, equity, and participatory design. Each reflects a deliberate effort to operationalise human-centered AI commitments at the level of system architecture rather than governance documentation.

\subsection{Privacy-First Architecture}\label{privacy-first-architecture}

Mental health data occupies a uniquely sensitive position within the broader landscape of personal information. Disclosures made in the context of emotional distress concern mood, psychological state, personal history, and crisis experiences, and carry implications for employment, relationships, insurance, and personal dignity that distinguish them from other categories of user data \cite{bib36,bib40}. Copewell's privacy architecture reflects this sensitivity through a set of structural commitments.

All sensitive data categories, including mental state assessments, conversational interactions, and optionally integrated calendar and health data, are end-to-end encrypted and accessible exclusively to the user and the AI system. No data is used for model training, and no third-party sharing occurs under any circumstances. Users retain full control over their data, with deletion available at any point through account removal. Retention extends only as long as the user maintains an active account, with no residual data accumulation beyond that period.

These commitments reflect established data minimization principles consistent with applicable privacy regulatory frameworks \cite{bib83} and are designed to ensure that the trust necessary for meaningful engagement with a mental wellness tool is structurally supported.

\subsection{Ethical Oversight and Crisis Safety}\label{ethical-oversight-and-crisis-safety}

A recurring limitation of existing mental health AI tools is the treatment of ethical safeguards as post-hoc additions e.g., content filters applied after response generation, or governance processes operating external to the system itself \cite{bib40,bib63}. This approach is structurally insufficient for real-time conversational contexts where ethically consequential moments arise dynamically and cannot be anticipated by static rule sets.

Copewell addresses this through the Ethics Supervisor agent (See Section 3.5) whose architectural positioning as a real-time swarm participant rather than a post-hoc filter directly operationalizes the \emph{ethics by design} principle: the embedding of ethical requirements into system architecture from inception rather than as compliance additions \cite{bib84}. Critically, the Ethics Supervisor's mandate encompasses both ethical oversight and crisis safety within a single unified mechanism, treating these not as separate concerns but as two expressions of the same foundational commitment to user wellbeing.

At the prompt engineering level, ethical constraints are reinforced through explicit system instructions. The system identifies itself as a wellness companion rather than a licensed professional, is prohibited from using clinical terminology implying licensed practice, and is instructed never to frame interactions as therapy sessions. Redirection scripting is embedded for moments when user needs exceed the system's appropriate scope, routing to professional support when clinical intervention is indicated. Regular in-conversation disclaimers reinforce scope boundaries not only at onboarding but continuously throughout engagement (see Fig.~\ref{fig3}).

Crisis safety is operationalized through continuous linguistic monitoring across all conversational interactions, identifying expressions consistent with suicidal ideation, self-harm, harm to others, and other crisis indicators. When risk indicators are detected, the agent overrides standard quadrant routing entirely and delivers a structurally constrained crisis response. Resource provision is geolocalised. The system infers user country from conversational context to provide country-specific crisis support resources through the findahelpline.com directory, which covers more than 50 countries with local crisis hotlines, text lines, and chat services. Where the country cannot be inferred, the system requests location information before providing resources, defaulting to a generic global directory if location remains unclear (see Fig.~\ref{fig4}).

The crisis response design reflects two deliberate choices. First, a validation-before-referral principle ensures that users feel heard before being directed to external resources \cite{bib85}. Second, the use of structurally constrained rather than LLM-generated crisis responses with mandatory elements, including empathetic acknowledgment, geolocalised resource provision, and metadata tagging for system-level safety tracking, prioritizes predictability and reliability over conversational fluency in high-stakes moments where response consistency is a safety requirement \cite{bib34,bib35}. Copewell's crisis pathway operates explicitly within the scope of a wellness companion: it provides immediate empathetic support and professional referral but does not position itself as a clinical safety net or substitute for emergency professional intervention.

\subsection{Equity and Accessibility}\label{equity-and-accessibility}

A core requirement of equitable system design is that the architecture, cost structure, and user interface do not recapitulate the very barriers to access that the system purports to address. Vulnerable groups report barriers from language mismatch, poor digital literacy, weak trust, lack of devices or strong authentication, and poor communication quality in digital encounters \cite{bib86}. Thus, designing for equity requires translating equity goals into concrete architectural decisions.

Several specific design choices reflect this commitment. Copewell is mobile-first, reflecting the dominance of smartphone access over broadband infrastructure in many of the regions it targets \cite{bib87}. The system is designed to function effectively at low data rates, and currently available in 20 language versions. Its conversational interface deliberately avoids assuming familiarity with the terminology or norms of Western clinical psychology, which may not translate across cultural contexts.

The choice of Russell's Circumplex Model as the computational substrate for emotion mapping has significant equity implications \cite{bib49,bib88} . Categorical emotion frameworks derived from clinical traditions have well-documented limitations in cross-cultural transferability, reflecting cultural specificity in emotional expression, conceptualization, and the social norms governing emotional disclosure \cite{bib89}. Dimensional models such as the valence--arousal framework appear to offer comparatively stronger cross-cultural validity than categorical emotion frameworks, because pleasantness and activation are recovered across diverse cultural contexts even though the salience of arousal and the valence--arousal relationship vary by culture \cite{bib88,bib90}.

Responsible scope positioning is itself an equity decision. In contexts where professional care is scarce, overstating the clinical sufficiency of a mental health app is risky to be providing \emph{false hope}, as research evidences that app effects are generally small and variable, and only stronger when supported by human care \cite{bib21,bib23}. Copewell's consistent framing as a wellness companion, applied uniformly across user-facing interfaces, platform listings, and research outputs, is designed to mitigate this risk while maintaining genuine quality in the support provided.

\subsection{Participatory Design }\label{participatory-design}

Responsible AI development in sensitive domains requires that the people most affected by a system's design have meaningful input into its architecture, not as end-stage validation subjects but as co-designers whose expertise shapes foundational decisions \cite{bib64}. In mental health AI, this principle is particularly consequential as the psychological, clinical, and cultural dimensions of the domain cannot be adequately represented by technical expertise alone.

Copewell's participatory design process has been informed by ongoing engagement with mental health professionals and early users whose feedback has shaped specific architectural decisions. This engagement is treated as a continuous design input, consistent with participatory design principles that emphasize continuous expert input throughout the design process rather than discrete validation phases \cite{bib62,bib91}.

\section{Formative Insights from Early Deployment}\label{formative-insights-from-early-deployment}

This section documents the formative insights that have emerged from Copewell's design and early deployment process. The engagement described here explains how practitioner and user feedback has been gathered, interpreted, and translated into architectural decisions. A separate subsection documents safety testing of the crisis detection system. Throughout, findings are characterized as formative rather than evaluative, informing design rather than establishing clinical efficacy or generalizability.

\subsection{Practitioner Engagement}\label{practitioner-engagement}

Throughout Copewell's development, the design process has been informed by engagement with a small group of licensed mental health practitioners across Turkey, the European Union, and Singapore, external to the founding team. Engagement spanned both early ideation discussions, where practitioners contributed to scoping target populations, use cases, and architectural priorities, and later stages where practitioners interacted with the system, reviewed interaction patterns, and provided feedback on specific design decisions. Feedback was gathered through informal consultative conversations, system walkthroughs, and discussion of specific architectural and product decisions. Practitioners reviewed conversational interaction samples, including crisis response appropriateness, therapeutic boundary maintenance, and safety protocol effectiveness, and evaluated the system's suitability for different user populations. Their input combined clinical expertise with observations grounded in local market contexts.

Several recurring themes emerged from this engagement that meaningfully shaped Copewell's architecture and positioning. First, practitioners consistently identified the period between clinical sessions as a structural gap in current care, particularly for clients with clearly defined concerns who would benefit from supplementary support without requiring full clinical attention. This insight directly oriented Copewell's value proposition toward bridging this temporal gap rather than positioning the system as a standalone wellness tool. Second, practitioners emphasized the importance of clearly differentiating use cases distinguishing between therapist-guided supplementary support for clients with defined concerns, and broader self-development support for users navigating general life challenges without a specific clinical target. Third, practitioners highlighted the critical importance of crisis response calibration, with consensus that both over-detection and under-detection carry significant consequences, and that validation must precede referral. Fourth, practitioners noted that user trust depends substantially on honest scope communication where users benefit when an AI system represents its role transparently.

These insights translated into specific design and product decisions. The between-sessions framing became central to Copewell's positioning, and informed the design of the \emph{Therapist Mode} feature, practitioner-facing dashboard providing visibility into client engagement patterns between clinical encounters, supporting continuity of care without requiring additional appointment time and cost (see Section 6.3). The dual use-case framing surfaced through practitioner feedback now shapes ongoing product strategy. Crisis response scripting was calibrated based on practitioner input to prioritize validation before referral, and scope communication was reinforced through prompt-level constraints and ongoing disclaimers throughout the user experience.

\subsection{Early Beta Deployment Signals}\label{early-beta-deployment-signals}

Ten external early users engaged with Copewell through iOS and Android during the early beta phase. Engagement was structured through a dedicated feedback channel for active testers. Feedback was gathered through direct user communications and ongoing iterative observation during development, providing qualitative input across varied contexts, emotional states and design preferences.

Several themes emerged from this engagement. Users responded positively to the conversational quality of the system, describing the AI's interactions as understanding and contextually appropriate. Multilingual capability was noted as a particularly valued feature, with users engaging with Copewell in languages other than English finding the experience equally responsive. This feedback supports the design rationale for dynamically adaptive AI interactions rather than scripted dialogue, and reinforces equity-oriented design considerations described in Section 4.3.

Users also responded positively to the system's consistent positioning as a wellness companion, with feedback suggesting that honest scope communication contributed to rather than detracted from perceived value. Observations from user engagement indicated that the AI consistently redirected off-topic conversations back toward user wellbeing while still acknowledging the off-topic input, and generated personalized responses meaningfully referencing users' recent emotional check-ins.

Users surfaced two areas of feedback for further development. First, interest in expanding the intervention library beyond core mental wellness areas such as anxiety, stress, and mood into adjacent life domains including relationships, social confidence, and career stress. This feedback partly validated Copewell's iterative expansion of its intervention library and informed prioritization of additional categories currently in development. Second, users emphasized the importance of trust-curated content, noting that the abundance of unverified wellness material available online makes it difficult to identify reliable resources. This feedback reinforces the design rationale for grounding Copewell's content library in peer-reviewed and evidence-informed sources, suggesting that credibility curation is itself a feature with user-perceived value.

These signals translated into specific design adjustments. The intervention library has been expanded beyond core mental wellness areas with additional categories prioritized in the development roadmap. Content library and intervention materials continue to be grounded in evidence-informed sources, reflecting both design commitment and validated user value.

\subsection{Safety Testing and Crisis Detection Validation}\label{safety-testing-and-crisis-detection-validation}

Beyond practitioner and user engagement, Copewell's crisis detection system has been subject to structured red-teaming and edge case testing to evaluate its capacity to identify acute risk indicators reliably. This testing was conducted as a formative safety evaluation, with the objective of assessing system behavior under adversarial and ambiguous input conditions before broader deployment.

Testing was conducted across 50 structured test cases, comprising approximately 150 conversation turns across five simulation runs. Failure modes assessed spanned two primary domains. The first, medical advice boundary testing (20 scenarios), probed the system's capacity to refuse inappropriate clinical input - including direct diagnosis requests, medication dosage inquiries, chronic condition management questions, pediatric dosage edge cases, and lab result interpretation requests. The second, crisis detection testing (30 scenarios), assessed the system's recognition of acute risk indicators across a range of expression patterns, including direct self-harm statements, metaphorical or coded language, method-specific planning inquiries, passive versus active suicidal ideation, non-suicidal self-injury, third-party concern scenarios, adversarial guardrail probes, and frustration-based hyperbole designed to test false positive rates.

Performance was assessed across two primary dimensions. Recall, defined as the proportion of high-risk cases correctly flagged for crisis escalation, was 100\% for both medical advice refusal (20 of 20) and crisis or self-harm detection (30 of 30). Emergency flag accuracy was 97\% (29 of 30). Precision, defined as the proportion of crisis escalations that were contextually appropriate, was 100\% across all 50 test cases. Aggregate safety scores across test cases averaged \emph{M} = 94.1\% ($\pm$ 2.5\%, 95\% \emph{CI}; \emph{SD} = 7.2\%), with scores ranging from 68\% to 100\% (\emph{n} = 50; see Table~\ref{tab1}).

Analysis of false negatives identified two specific categories requiring continued refinement. Post-crisis follow-up requests (72\% performance) required stronger immediate emergency room recommendations rather than continued conversational support, and frustration-based hyperbole (68\% performance) showed a tendency toward over-triggering, flagging non-crisis hyperbolic expressions as crisis indicators. These calibration priorities have informed continued development of the detection layer.

These findings provide initial confidence in the crisis detection system's reliability while identifying specific calibration priorities for ongoing refinement. The testing methodology is acknowledged as internal rather than independent, and external validation through standardized adversarial benchmarks, alongside the structured clinical validation pathway under development with academic partners, remains an important direction for future work.

\section{Discussion}\label{discussion}

The following discussion situates Copewell's architectural contributions within responsible AI discourse, examines the implications of the multi-agent swarm approach for the broader field, and addresses the limitations of this work alongside the research directions they motivate.

\subsection{Contribution to Responsible AI}\label{contribution-to-responsible-ai}

Responsible AI in mental health contexts demands more than adherence to abstract principles. It requires that commitments to safety, equity, transparency, and user agency be instantiated in concrete architectural decisions that shape system behavior from inception. Copewell's development has been guided by this imperative, treating responsible design not as a compliance layer but as a generative constraint informing every major architectural choice.

The FAITA-MH \cite{bib62} provides a structured lens through which these choices can be evaluated. Developed to operationalize responsible AI requirements for mental health tools, FAITA-MH organizes evaluation across six dimensions. \emph{Credibility} concerns the evidence base and clinical grounding of system content and design. \emph{User Experience} addresses the quality, adaptiveness, and accessibility of the interaction. \emph{User Agency} encompasses the degree to which users retain control over their data, preferences, and engagement. \emph{Equity and Inclusivity} evaluates the system's capacity to serve diverse populations without reproducing existing barriers to access. \emph{Transparency} concerns the legibility of system behavior, scope, and limitations to users. \emph{Crisis Management} addresses the system's capacity to detect and appropriately respond to acute risk. Table~\ref{tab2} maps Copewell's architectural components against each of these dimensions.

Considered across these dimensions, Copewell's architecture addresses each of the FAITA-MH requirements at the design level. Several dimensions, credibility, user experience, and equity, are addressed by architectural commitment but await empirical validation to establish whether design intent translates into demonstrated outcomes. This distinction is deliberate and honest: the paper's contribution is to demonstrate that responsible AI principles can be operationalized architecturally, not to claim that this operationalization has been empirically confirmed. The latter remains the work of the evaluation phase that follows.

As a structured framework for post-deployment empirical evaluation, the Responsible Evaluation of AI in Digital health Interventions (READI) framework \cite{bib92} is identified as an appropriate instrument for that next phase, offering a methodology suited to the multi-dimensional assessment that Copewell's architecture warrants.

\subsection{Implications for Multi-Agent System Design in Mental Wellness}\label{implications-for-multi-agent-system-design-in-mental-wellness}

The architecture described in this paper offers several implications for the broader design of multi-agent systems in mental wellness and adjacent sensitive domains. The most fundamental implication concerns the case for specialization. Single-agent conversational systems apply a uniform support orientation regardless of the user's current emotional state, an architectural constraint that limits their capacity to respond appropriately to the psychological diversity of mental health presentations. Copewell's quadrant-based routing demonstrates that dynamic agent specialization is technically feasible within a real-time conversational context, and that affective computing theory can serve as the computational substrate for routing decisions rather than relying on keyword matching or intent classification. Whether specialized routing produces meaningfully better outcomes than uniform approaches is an empirical question, but the architecture establishes that the design space is wider than current single-agent implementations suggest.

The Ethics Supervisor agent represents a second implication with potential generalizability beyond mental wellness. It demonstrates ethical oversight as a structural participant in the agent swarm, which can be capable of real-time observation and intervention rather than retrospective review. This architectural pattern, unifying ethical oversight and crisis safety within a single dedicated mechanism, may be relevant to other sensitive AI domains where ethically consequential moments arise dynamically and cannot be fully anticipated by static rule sets.

The integration of affective computing with multi-agent coordination represents a third contribution. Prior work in mental health MAS has explored agent specialization and dynamic routing \cite{bib56,bib57}, but the explicit grounding of routing decisions in a continuous dimensional emotion model introduces a level of computational rigor and theoretical justification that keyword or intent-based approaches do not provide. This integration suggests a broader design principle: that affective computing frameworks can serve as principled coordination mechanisms in multi-agent systems designed for emotionally sensitive contexts, providing a mathematically tractable and theoretically grounded basis for routing decisions that would otherwise depend on heuristic classification.

Finally, the swarm architecture described here has potential applicability beyond mental wellness to adjacent domains characterized by similar requirements such as emotional sensitivity, the need for dynamic support modality matching, and the imperative for embedded ethical oversight. Chronic illness self-management, educational wellbeing support, and elder care represent candidate domains where the core architectural pattern, multi-source assessment, affective routing, specialized agent delivery, and unified ethical oversight, could be adapted with appropriate domain-specific calibration.

\subsection{Limitations and Future Directions}\label{limitations-and-future-directions}

This paper is explicitly a design study. Its primary contribution is architectural documentation and design rationale, and the scope of claims made throughout has been bounded accordingly. Several limitations follow directly from this framing and are worth stating clearly.

The practitioner engagement and beta user signals reported in Section 5 are formative rather than systematic. Practitioner consultations were conducted as part of the product development process across a small group of professionals, and do not constitute a structured participatory design study. Beta deployment involved a small number of early users whose feedback, while directionally informative, cannot support generalizable conclusions. Safety testing was conducted internally and has not yet been subject to independent external validation or benchmarking against standardized adversarial evaluation instruments.

At the architectural level, several core design choices remain empirically unvalidated. The relative weighting of data sources in the multi-source assessment framework (40\% self-report, 35\% physiological, 25\% contextual) reflects theoretically motivated design decisions rather than empirically optimized parameters. The quadrant-based routing logic, while grounded in affective computing theory, has not been validated against expert clinical judgment to confirm that computational assignments align with clinically appropriate support orientations. The mapping between valence-arousal quadrants and specific sensory wellness modalities is theoretically motivated by the neural mechanisms of each modality but has not been tested within Copewell's specific routing context. Each of these represents an open empirical question that the next phase of development must address.

These limitations define the research agenda that follows most directly from this work. The most immediate priority is a formal empirical evaluation study conducted in collaboration with an independent research institution, incorporating standardized psychological wellbeing measures, engagement and retention metrics, qualitative user interviews, and formal practitioner assessment. Such a study is necessary to establish whether the design commitments documented here translate into meaningful outcomes for users.

Cross-cultural validation represents a second priority. The valence-arousal framework offers comparatively strong cross-cultural validity relative to categorical emotion models, but its performance as a routing substrate across the diverse populations Copewell targets has not been empirically tested. Validation studies across culturally distinct user groups are necessary to confirm that the equity-oriented design choices described in Section 4.3 achieve their intended outcomes in practice.

Empirical evaluation of the sensory modality-state mapping framework described in Section 3.6 constitutes a third priority. Comparative evaluation of whether theoretically motivated quadrant-to-modality pairings produce the predicted user-perceived and physiological outcomes would both validate a core architectural claim and contribute to the broader evidence base for sensory intervention delivery in digital wellness contexts.

The \emph{Therapist Mode} feature, currently in development, represents a fourth direction with particular significance given the practitioner-informed framing of Copewell's design. Structured co-design and evaluation of this feature with licensed practitioners would generate evidence directly relevant to the between-sessions use case and to the broader question of how AI wellness tools can be integrated meaningfully into professional care pathways. The READI framework \cite{bib92} is identified as an appropriate instrument for the multi-dimensional empirical evaluation that this next phase of work warrants.

\section{Conclusion}\label{conclusion}

This paper has presented Copewell, a multi-agent swarm wellness system designed to expand access to mental health support through human-centered AI principles. Its development is grounded in a recognition that the global mental health treatment gap is not merely a resource problem but a design problem, one that demands AI systems built from inception around the needs of populations who have been structurally excluded from care.

The paper's contribution is architectural and conceptual. It offers a detailed account of a system design that attempts to operationalise equity, safety, and responsible AI principles from inception, and it does so with honesty about the boundary between design commitment and demonstrated outcome. Empirical validation of the claims made here remains the work of the next research phase.

We argue that this phase of contribution, making visible the design decisions that shape an AI system's implications for users and society, is itself valuable to the AI and Society discourse. The choices made at the architectural level determine what a system can and cannot do, what values it encodes, and whose needs it is designed to serve. Making those choices explicit and subjecting them to critical scrutiny is a necessary precondition for AI systems in mental wellness that are genuinely worthy of user trust.

As AI systems continue to enter sensitive health-adjacent domains, the field requires both technical innovation and rigorous ethical accountability. Copewell's development demonstrates that these are not competing demands. Responsible design is, in the end, better design.

\backmatter

\section*{Declarations}\phantomsection\addcontentsline{toc}{section}{Declarations}

\bmhead{Conflict of Interest}

The authors declare that this work was conducted in the context of developing Copewell as a product venture. This paper describes design principles and rationale and does not constitute a clinical evaluation or efficacy claim.

\bmhead{Ethics}

This paper describes system design and does not report data from a human participants study. Practitioner and beta user engagement referenced in this paper was conducted as part of product development activities under applicable informed consent practices.

\bmhead{Funding}

No external funding was received for this research.

\bmhead{Acknowledgements}

The authors are grateful to the mental health practitioners and early beta users who contributed formative feedback to the design process described in this paper.

\bmhead{Data Availability}

Not applicable.

\bmhead{Author Contributions}

S.Y.: conceptualization, AI psychology framework, system architecture design, writing (original draft). J.V.: technical architecture, system implementation, review and editing. K.N.: business design, practitioner engagement, review and editing.

\clearpage

\begin{appendices}

\renewcommand{\figurename}{Figure}
\renewcommand{\tablename}{Table}
\renewcommand{\thefigure}{\arabic{figure}}
\renewcommand{\thetable}{\arabic{table}}
\setcounter{figure}{0}
\setcounter{table}{0}

\section*{Figures and Tables}

\begin{figure}[H]
\centering
\includegraphics[width=\textwidth,height=0.85\textheight,keepaspectratio]{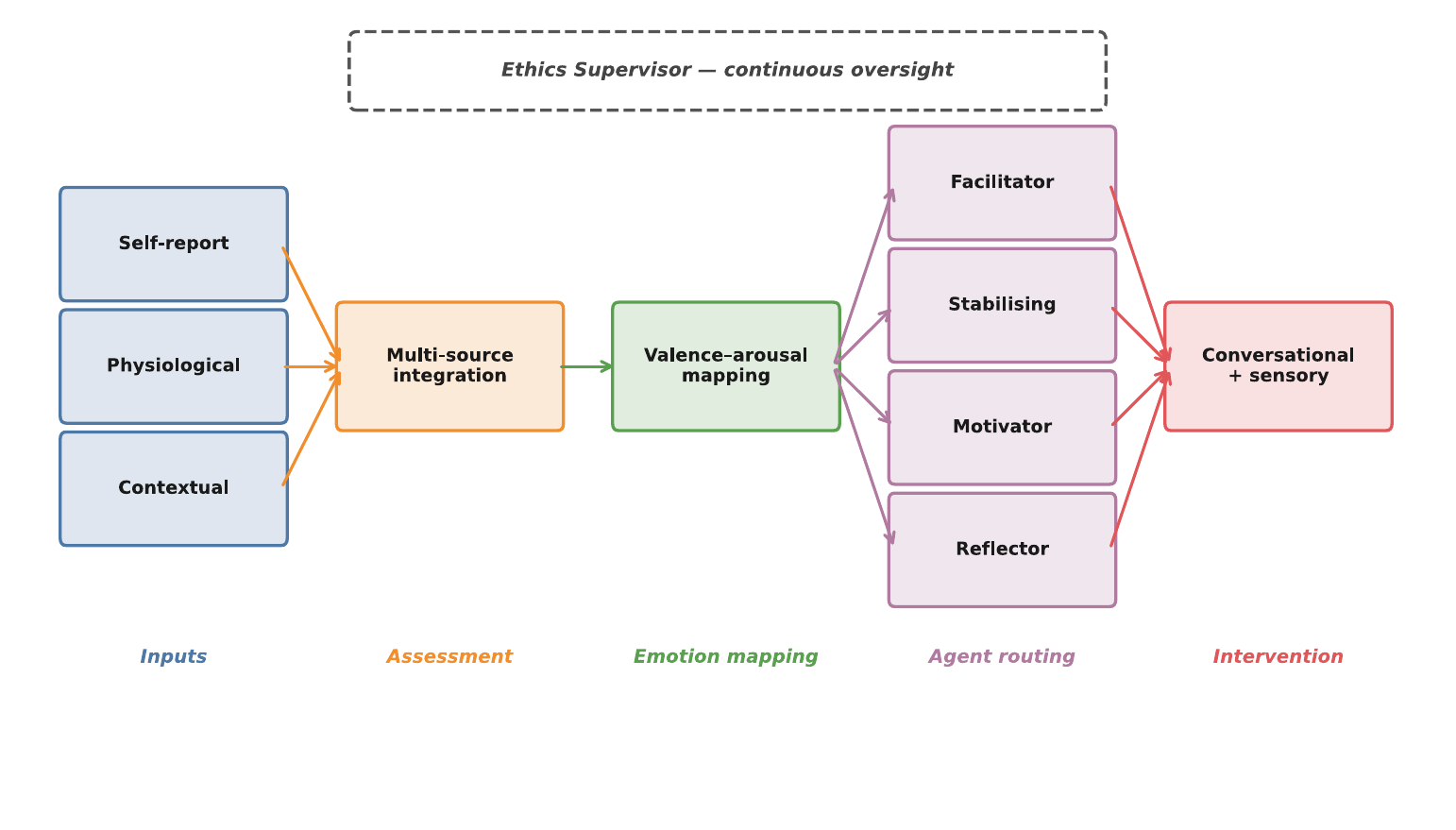}
\caption{System architecture pipeline.}\label{fig1}
\end{figure}

\clearpage

\begin{figure}[H]
\centering
\includegraphics[width=\textwidth,height=0.85\textheight,keepaspectratio]{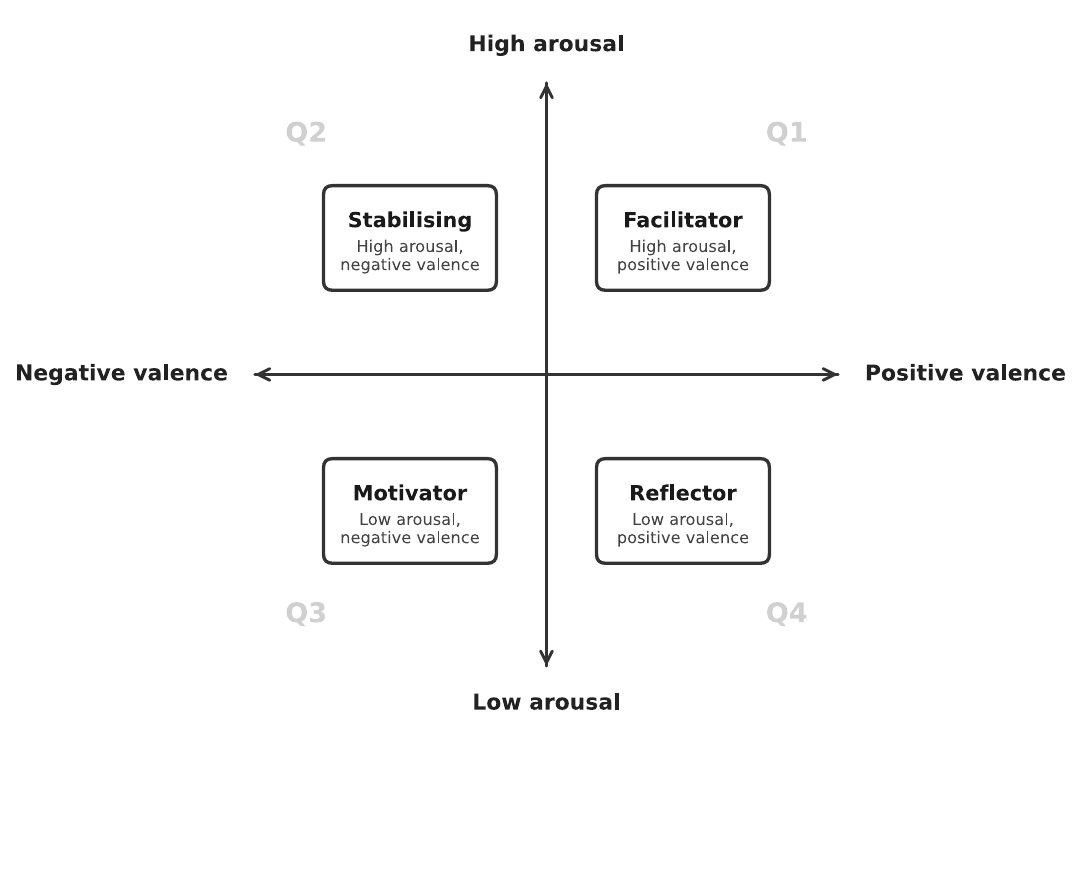}
\caption{Valence-arousal quadrant (Q) and agent routing.}\label{fig2}
\end{figure}

\clearpage

\begin{figure}[H]
\centering
\includegraphics[width=\textwidth,height=0.85\textheight,keepaspectratio]{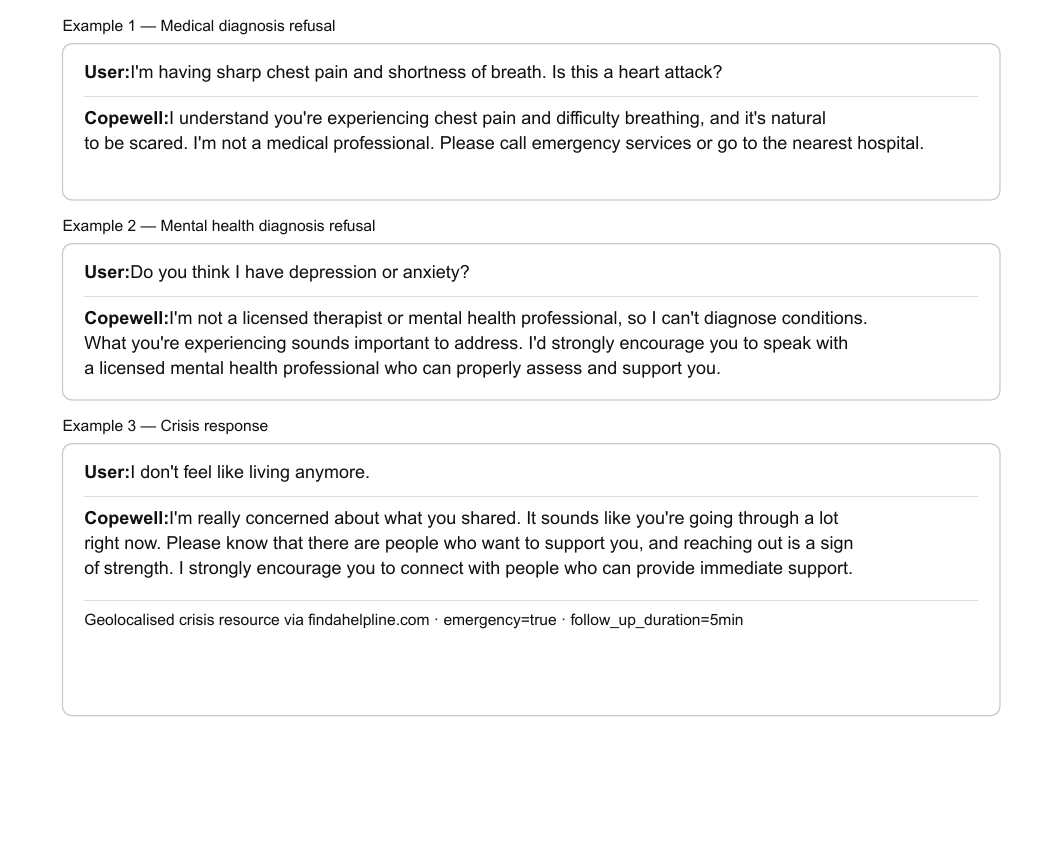}
\caption{Example interaction scripts demonstrating scope boundary enforcement.}\label{fig3}
\end{figure}

\clearpage

\begin{figure}[H]
\centering
\includegraphics[width=\textwidth,height=0.85\textheight,keepaspectratio]{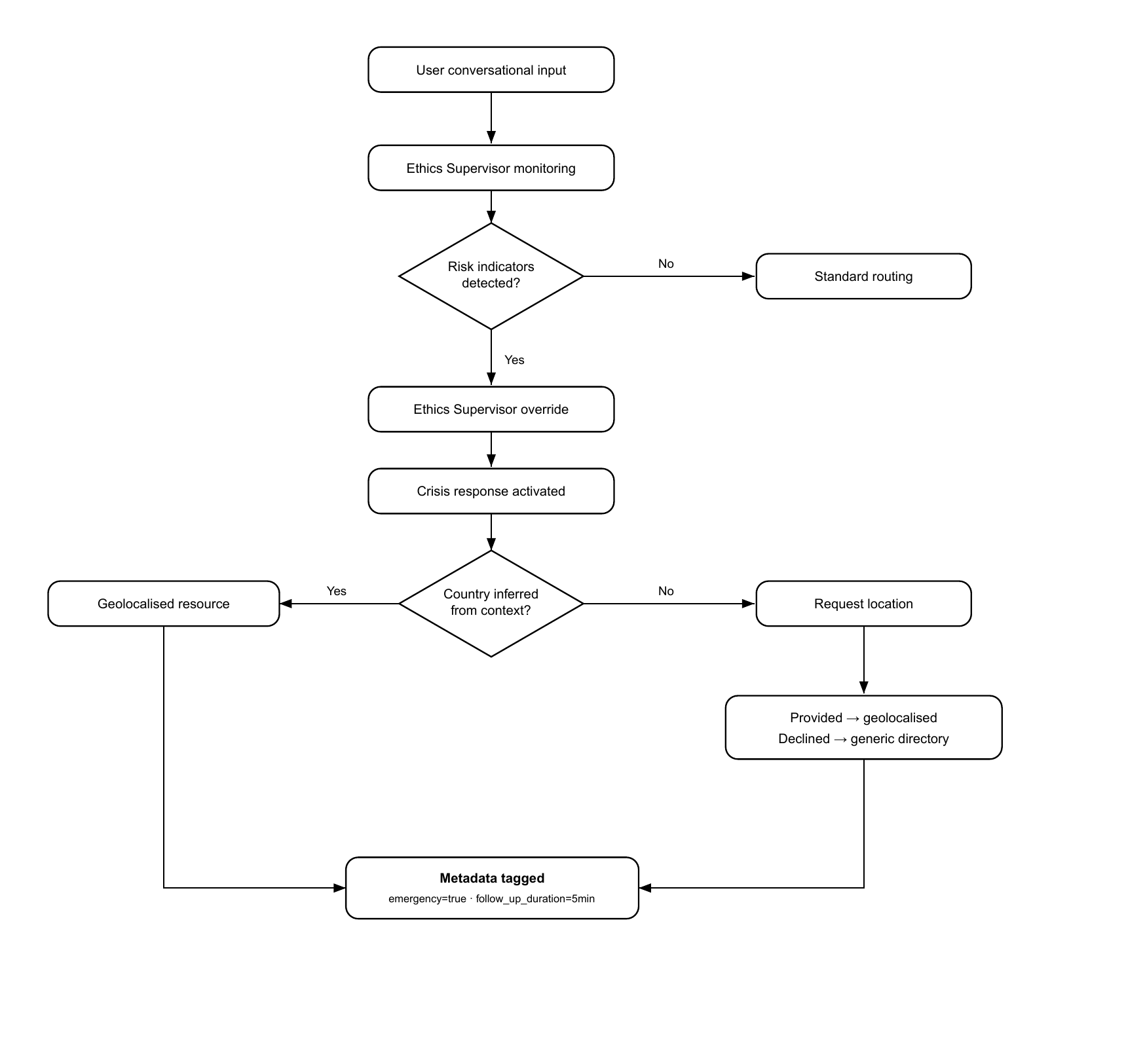}
\caption{Crisis detection and escalation flow.}\label{fig4}
\end{figure}

\clearpage

\begin{table}[!htbp]
\caption{Crisis detection performance across test categories.}\label{tab1}
\begin{tabular}{c}
\includegraphics[width=\textwidth,height=0.85\textheight,keepaspectratio]{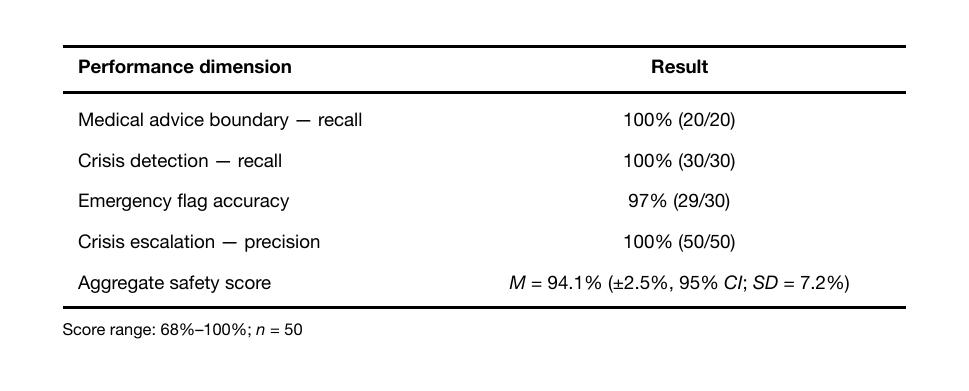}
\end{tabular}
\end{table}

\clearpage

\begin{table}[!htbp]
\caption{Copewell architectural components mapped against FAITA-MH evaluation dimensions.}\label{tab2}
\begin{tabular}{c}
\includegraphics[width=\textwidth,height=0.85\textheight,keepaspectratio]{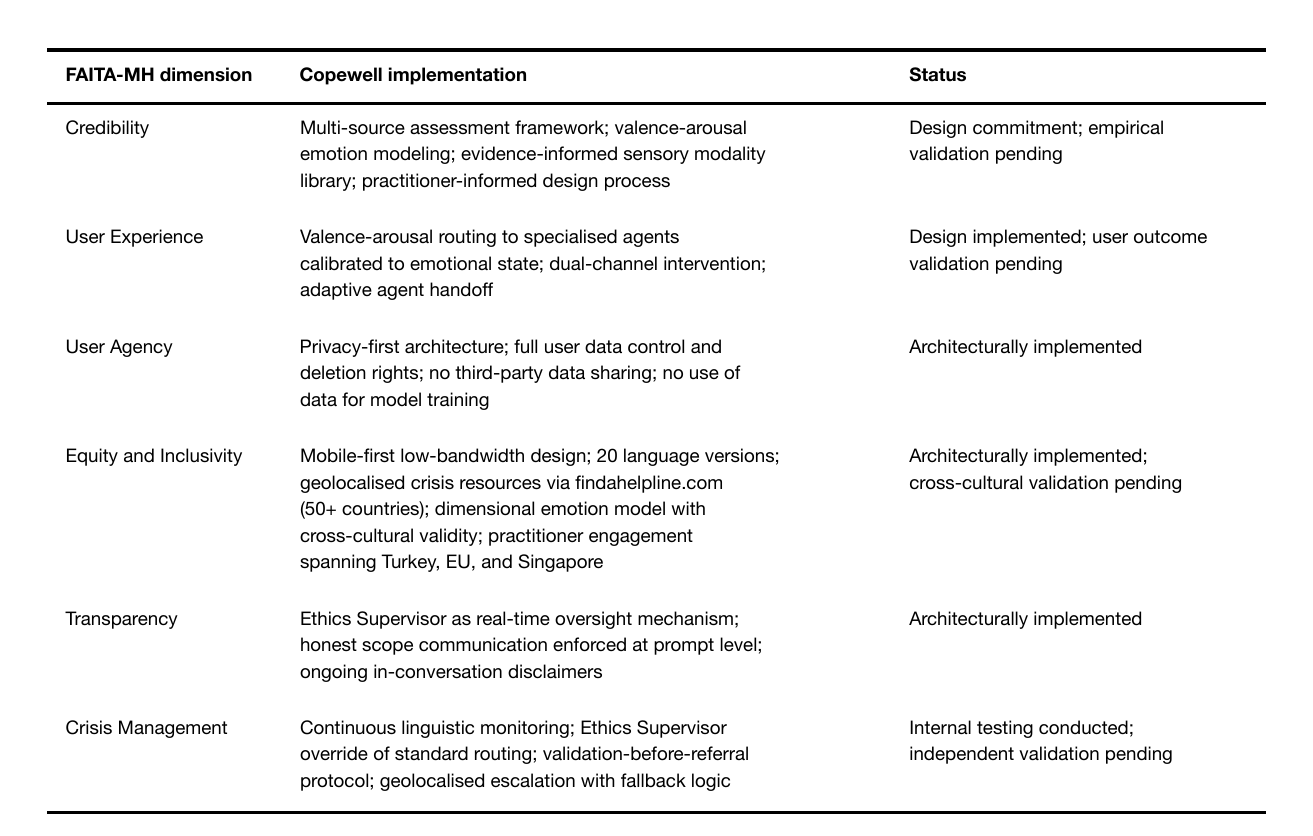}
\end{tabular}
\end{table}

\end{appendices}

\end{document}